%% file: main.tex
\pdfoutput=1

\documentclass[11pt]{article}

\usepackage[preprint]{acl}

\usepackage{times}
\usepackage{latexsym}

\usepackage[T1]{fontenc}

\usepackage[utf8]{inputenc}

\usepackage{microtype}

\usepackage{inconsolata}

\usepackage{graphicx}

\usepackage{listings}
\usepackage{nicefrac}       
\usepackage{multirow} 
\usepackage{booktabs}
\usepackage[most]{tcolorbox}
\usepackage{enumerate}
\newtcolorbox{promptbox}[1][]{
  breakable,
  colframe=blue!30!white,    
  colback=blue!10!white,     
  coltitle=black,            
  title=#1,                  
  rounded corners,           
  boxrule=0.5mm,             
  boxsep=5pt,                
  toptitle=1mm,              
  bottomtitle=0.5mm,           
  left=10pt,                 
  right=10pt,                
  top=5pt,                   
  bottom=5pt,                
  fonttitle=\bfseries        
}
\usepackage{enumitem}
\newlist{hyphenitemize}{itemize}{1}
\setlist[hyphenitemize,1]{label=-}
\usepackage{bbding}

\title{Bottom-Up Synthesis of Knowledge-Grounded Task-Oriented Dialogues with Iteratively Self-Refined Prompts}

\author{
 \textbf{Kun Qian\textsuperscript{1}},
 \textbf{Maximillian Chen\textsuperscript{1}},
 \textbf{Siyan Li\textsuperscript{1}},
 \textbf{Arpit Sharma\textsuperscript{2}},
  \textbf{Zhou Yu\textsuperscript{1,3}}
\\
 \textsuperscript{1}Department of Computer Science, Columbia University\\
 \textsuperscript{2}Walmart
 \textsuperscript{3}Arklex.ai
 }

\newcommand{\methodnospace}{\textit{BUSY}}
\newcommand{\method}{\textit{BUSY }}
\begin{document}
\maketitle
\begin{abstract}

Training conversational question-answering (QA) systems demands a substantial amount of in-domain data, which is often scarce in practice. A common solution to this challenge is to generate synthetic data. Traditional methods typically follow a top-down approach, where a large language model (LLM) generates multi-turn dialogues from a broad prompt. While this method produces coherent conversations, it offers limited fine-grained control over the content and is susceptible to hallucinations. 
We introduce a bottom-up conversation synthesis approach, where QA pairs are generated first and then combined into a coherent dialogue. This method offers greater control and precision by dividing the process into two distinct steps, enabling refined instructions and validations to be handled separately. Additionally, this structure allows the use of non-local models in stages that do not involve proprietary knowledge, enhancing the overall quality of the generated data. Both human and automated evaluations demonstrate that our approach produces more realistic and higher-quality dialogues compared to top-down methods.

\end{abstract}

\input{Content/Introduction}
\input{Content/Methods}
\input{Content/Eval}
\input{Content/Conclusion}
\section{Acknowledgement}
We express our sincere acknowledgment to Walmart for their support throughout this project. The access to real-user data provided by Walmart is inspiring for synthesizing conversational QA data of high quality and diversity. We are also grateful for the collaboration and insights shared by Walmart’s team, which greatly enhanced the depth and relevance of our work. 
\input{Content/Limitation}

\bibliography{anthology,custom}

\appendix
\onecolumn
\input{Content/Appendix_datasets}
\newpage

\input{Content/Appendix_qa_eval}

\input{Content/Appendix_dialogue_principles}
\newpage
\input{Content/Appendix_QualitativeAnalysis}
\newpage
\input{Content/Appendix_prompt_question}
\input{Content/Appendix_prompt_answer}
\input{Content/Appendix_prompt_attr}
\input{Content/Appendix_prompt_dialog}
\input{Content/Appendix_human_eval}
\input{Content/Appendix_dialogue_eval}

\end{document}

%% file: Content/Introduction.tex
\section{Introduction and Related Work}
Acquiring high-quality, in-distribution data is always the major challenge in building deployable conversational assistants. To address this challenge, researchers have developed more sample-efficient training methods for generative models. These methods include dialogue-specific pre-training objectives \cite{he2022galaxy}, improved domain adaptation through embedding learning \cite{zhao2018zero} and meta-learning \cite{qian2019domain}, or reinforcement learning approaches for task-oriented dialogues \cite{chen2024learning}.
However, such training methods are either computationally expensive or still rely on having sufficient cross-domain or in-domain seed data.

An increasingly popular strategy is to leverage large language models (LLMs) to 
synthesize dialogue data~\cite{chen-etal-2023-places,kim-etal-2023-soda}. Such methodologies for generating conversational data predominantly adopt \textbf{a top-down approach}: 
given a high-level outline, an LLM is typically asked to synthesize complex multi-turn interactions in a single pass. While this approach can produce coherent dialogues, it often lacks the granularity necessary for creating nuanced and realistic conversational datasets~\cite{zhou2024real, hayati2023far}, as the instruction is long and sometimes LLM will ignore some aspects of the instruction. This is especially true in the virtual assistant setting, where conversations often emphasize question-answering to fulfill information-seeking or task-oriented requests and not social interaction.
In addition, when dialogue generation relies on external knowledge, top-down approaches often require access to databases, raising privacy concerns when using non-local LLM models.

To improve conversation synthesis in such task-oriented and knowledge-grounded settings, we propose \textbf{B}ottom-\textbf{U}p Conversation \textbf{Sy}nthesis (BUSY).
Our \textbf{bottom-up} framework for dialogue dataset construction begins with generating high-quality question-answer (QA) pairs, which serve as the foundation for grounding complex dialogues in factual information. These questions are iteratively refined through automatic improvements to large language model (LLM) prompts. The corresponding answers are generated using the product database, with an emphasis on factual accuracy over naturalness. To ensure privacy, a local model is employed for answer generation, maintaining the confidentiality of the database. Then, we integrate these QA pairs with introductory, concluding, and connecting dialogue turns to create coherent and contextually relevant conversations.


\begin{figure*}[t]
\centering
  \includegraphics[width=0.9\textwidth]{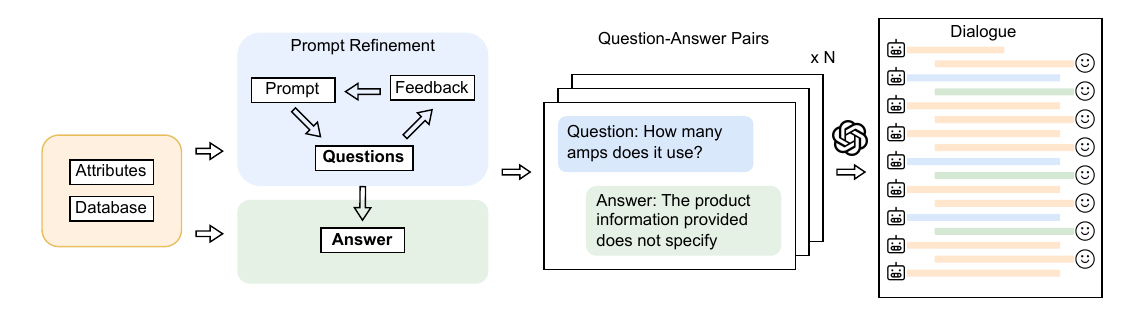}
  \caption{Framework for bottom-up dialogues synthesis. First, we iteratively refine the prompt to generate realistic questions by returning the comparison between generated questions and real-user question examples. Then, we prompt LLMs to generate an answer with the corresponding database information. We randomly sample N number of question-answer pairs and prompt LLMs to construct a dialogue by connecting these QA pairs.}
  \label{fig:framework}
\end{figure*}

We apply \method to the e-commerce domain~\cite{balakrishnan2024conversational, bernard2023mg,chiu2022salesbot}. These interactions are strictly task-oriented: assistants streamline various customer service processes (e.g., answering factual queries or guiding users through purchasing decisions), which can greatly improve consumers' shopping experience \cite{borges2010does,granbois1968improving}. Moreover, due to the monetary implications of conversations in this type of domain, all QA pairings must be grounded on factual knowledge verifiable by a knowledge base. Using our framework, we produce a synthetic corpus called the Shopping Companion Dialogues (ShopDial), which consists of 6,000 dialogues spanning several different shopping categories\footnote{Dataset and code is available at \url{https://github.com/qbetterk/ConvQA_Walmart}}. We employ human annotators and LLM agents to validate the quality of our synthetic dialogues. Our experimental results demonstrate that the use of iteratively self-refined prompts leads to realistic question generation, and the bottom-up synthesis framework effectively ensures the quality of the synthesized dialogues.


%% file: Content/Methods.tex
\section{\methodnospace: Bottom-Up Conversation Synthesis}
\looseness-1
Figure~\ref{fig:framework} describes our framework. We first generate pairs of domain-relevant questions and knowledge-grounded answers. Then, we connect these pairs into conversations.



\begin{table*}[t]
\small
\centering
\scalebox{0.9}{
\begin{tabular}{llccccccc}
\toprule
\multicolumn{2}{c}{Iteration:} & 1 & 2 & 3 & 4 & 5 & 6 \\ \midrule
\multirow{6}{*}{Question} & Human                          & 0.83 & 0.93 & \textbf{1.00}    & 0.99 & 0.97 & \textbf{1.00}    \\ \cmidrule(lr){2-8}
                          & Brand Safety (Q)                & 1.00 & 0.99 & 0.99 & 0.99 & 0.99 & 0.99 \\
                          & Brand Preference (Q) $\downarrow$ \footnotemark{}           & 0.89 & \textbf{0.81} & 0.85 & 0.83 & 0.82 & 0.78 \\
                          & Customer Safety (Q)      & 1.00 & 1.00 & 1.00 & 1.00 & 1.00 & 1.00 \\
                          & Friendliness (Q)                & 0.92 & \textbf{1.00} & \textbf{1.00} & 0.99 & 0.98 & 0.98 \\
                          & Quality (Q)                    & 0.58 & \textbf{0.82} & 0.77 & 0.80 & 0.75 & 0.76 \\ \midrule
\multirow{10}{*}{Answer}  & Brand Safety (A)           & 0.97 & 0.97 & 0.97 & 0.96 & 0.97 & 0.97 \\
                          & Brand Preference (A) $\downarrow$  & 0.99 & 0.97 & 0.98 & 0.98 & 0.97 & 0.98 \\
                          & Customer Safety (A)           & 1.00 & 1.00 & 0.99 & 1.00 & 1.00 & 1.00 \\
                          & Friendliness (A)        & 0.54 & \textbf{0.63} & 0.62 & 0.60 & 0.58 & \textbf{0.63} \\
                          & Quality (A)        & 0.54 & 0.57 & \textbf{0.58} & 0.55 & 0.55 & 0.56 \\
                          & Question Relevance (A)      & 0.93 & 0.89 & 0.90 & 0.88 & 0.89 & 0.92 \\
                          & Prompt Leakage (A)      & 1.00 & 1.00 & 1.00 & 1.00 & 1.00 & 1.00 \\
                          & Truthfulness (A)         & 0.95 & 0.94 & 0.96 & 0.95 & 0.96 & 0.96 \\
                          & Entailment (A)              & 0.99 & 0.99 & 0.98 & 1.00 & 0.98 & 0.98 \\ \bottomrule
\end{tabular}
}
\caption{Automatic and human evaluation of synthetic questions and answers on e-commerce metrics over different prompt-editing iterations. Our approach significantly improves data quality in terms of brand preference, friendliness, and overall quality, as well as human evaluation. The improvement converges after the third iteration.}
\label{tab:eval_qa}
\end{table*}

\subsection{Question Generation}
\looseness-1

To construct realistic, diverse, and accurate questions, we divide the task into three steps. First, we extract attributes from existing in-domain seed questions. We collect 20 human-written questions as seed questions for each domain. Second, we generate questions by iteratively refining LLM prompts. Finally, we validate that the generated questions contain the desired attributes.

In e-commerce and other related domains, customers ask factual questions about diverse entity attributes (e.g. a product's color, specifications, or reviews). To create a diverse yet knowledge-grounded question set, we need to mimic the question structure but create variations on these attribute types. Therefore, we first prompt LLMs to extract the attribute of each seed customer question (see Appendix~\ref{appd:attr}) and all possible attributes of each category from the product database. Then, we ask LLMs to select at most three of the most relevant attributes for each seed question.

After obtaining the attributes of the original seed questions, we prompt LLMs to generate new questions.
Similar to ~\citet{wan2023universal}, we use downstream task feedback to identify an ``optimal'' task-specific prompt to generate questions in a three-step approach: (1) We write a coarse prompt and ask LLMs to generate questions based on attributes. (2) We ask LLMs to compare the seed and the generated questions (see Appendix~\ref{appd:prompt_eval}), which share the same attributes. (3) Based on this comparison, we ask an LLM to edit the generation prompt. We repeat steps (2) and (3) until the prompt does not change 
(see Appendix~\ref{appd:gen_q} for an example of an initial prompt and an optimized prompt). In our experiments, the process terminated around six iterations (Sec.~\ref{sec:eval}). This fully automated process is simple and effective and results in quality improvements without significant prompt engineering. 

Previous prompt-based synthesis method stresses the importance of post-processing due to LLM-based generation not having hard constraints ~\cite{kim-etal-2023-soda, chen2022weakly}.
Similarly, we validate the synthesized questions by extracting attributes from the generated questions and ensuring they align with the attributes they were conditioned on. If the target attributes are not matched, we continue re-generating the questions until they meet the desired criteria or the maximum number of generation attempts is reached. Once the prompt is finalized, we use it to prompt the LLM to generate questions for all attributes in order to ensure diversity.

\begin{table*}[t]
\centering
\small
\scalebox{0.9}{
\begin{tabular}{lccccccc}
\toprule
        & \multicolumn{3}{c}{\textbf{LLM Eval}}      & & \multicolumn{3}{c}{\textbf{Human Eval}} \\ \midrule
        & PLACES & CoQA  & ShopDial  & & PLACES & CoQA  & ShopDial  \\ \midrule
Coherence    & 4.55   & 4.9  & \textbf{4.95} & & \textbf{4.15}   & 3.62 & 4.05 \\
Informativeness    & 3.55   & 3.65 & \textbf{3.95} & & \textbf{4.25}   & 3.85 & 3.78 \\
Truthfulness    & 4.55   & 4.01 & \textbf{4.70}  & & 4.25   & 4.17 & \textbf{4.48} \\
Naturalness    & 4.50    & \textbf{4.90}  & 4.85 & & 3.30    & 2.97 & \textbf{3.33} \\
Completeness    & 3.90    & 3.97 & \textbf{4.25} & & \textbf{4.18}  & 3.30  & 4.00 \\ 
Overall & 3.90    & 4.12 & \textbf{4.25} & & 3.59   & 3.17 & \textbf{3.63} \\ \bottomrule
\end{tabular}}
\caption{LLM-based dialogue evaluation (left) and human evaluation (right) in terms of scores in six metrics. }
\label{tab:dial_eval}
\end{table*}

\vspace{-0.1cm}

\subsection{Answer Construction with Database}
\looseness-1
\label{sec:gen_a}
We require our answers to be truthful, which means each answer is generated based on the attribute values from a database. Therefore, to answer each generated question, we sample a product from the database under the corresponding category first. Then, we extract the value of the relevant attributes of the question. We construct each question's answer based on the sampled attribute value. However, some products do not have complete values for each attribute. Following the notion of selective prediction~\cite{chen2023adaptation}, in these unanswerable cases, we use templates such as ``\textit{I'm sorry, but I don't have the specific information for ...}'' to prevent hallucination. As is common in industrial settings, the product information may be confidential in certain cases, so we strictly use locally deployable models such as Llama 3 Instruct~\cite{feng-etal-2024-llama} to generate answers. This is the only step in our entire synthesis pipeline where attribute values from the database are accessed.

\subsection{Connecting QA Pairs into Conversations}
\looseness-1
Once we have high-quality QA pairs, the next step, as indicated in Figure~\ref{fig:framework}, is to connect them into complete, coherent conversations by prompting LLMs (Appendix~\ref{appd:gen_dial}).
We apply our framework to the e-commerce domain. Our intended scenario involves a customer navigating a product page on an online retail site and interacting with a shopping companion. This companion is a virtual assistant integrated into the website with full access to product databases (see Appendix~\ref{appd:principles} for more generation details).

This process leads to the creation of the Shopping Companion Dialogues (ShopDial) dataset, which encompasses six categories: \textit{vacuums, diapers, sofas, TV, food, and clothing}. The database of categories provides more than 500 products, resulting in 1,000 dialogues per category with an average of 8.03 turns per dialogue. These turns contain at least three product-relevant question-answer pairs and, on average, 1.3 ``unknown'' turns.
Table~\ref{tab:dataset_comparison} (Appendix~\ref{appd:dataset}) compares our ShopDial and other dialogue datasets. We are the first to generate dialogues using a bottom-up approach, as well as to introduce a synthetic dialogue dataset specifically tailored to the e-commerce domain. Fig.~\ref{fig:dial} illustrates an example from our ShopDial. This example demonstrates that our framework effectively produces high-quality question-answer pairs while ensuring natural transitions between turns. The example dialogue also includes “unknown” turns, where the assistant lacks sufficient information to respond. There are also instances of negative feedback from users, mimicking real-life user sentiments. Incorporating these elements enhances the ability of virtual assistants trained with ShopDial to manage realistic scenarios effectively.

%% file: Content/Eval.tex
\section{Evaluation and Results}
\label{sec:eval}

\subsection{Question-Answer Pair Evaluation}
Table~\ref{tab:eval_qa} presents the scores from both human evaluation and automatic metrics from a large e-commerce retailer over different prompt-refinement iterations. Similar to human evaluation, each metric is presented as a multiple-choice question, and each choice represents a certain level of that metric. 
Due to space constraints, the metrics are described in detail in Appendix~\ref{appd:qa_eval}.  
We observe significant improvements in the scores for branch preference, friendliness, overall quality, and human evaluation after iterative modifications to the generation prompt. These enhancements are attributed to the targeted refinements in the prompt that specifically highlight these aspects. For instance, a guideline to avoid bias towards any unmentioned brands was incorporated into the prompt following the second iteration. Additionally, the improvements appeared to converge after the third iteration, indicating that our method of iterative self-refinement for prompt editing effectively identifies and addresses discrepancies between generated and example questions, leading to efficient, prompt modifications. For most other metrics, the synthetic data consistently achieved near-perfect scores across all iterations, underscoring the robustness of the generation model.

\subsection{Synthetic Dialogue Evaluation}
For dialogue evaluation, we compare our method with an established top-down dialogue generation framework, PLACES~\cite{chen-etal-2023-places}. Following their work, we use expert-filtered synthetic dialogues from ShopDial as the in-context dialogue examples, resulting in 200 new synthetic dialogues to be used for evaluation. We also compare ShopDial to synthetic dialogues generated using PLACES with random examples from CoQA~\cite{coqa}, a popular human-collected conversational QA dataset. To ensure a fair comparison in product relevance, we include the product database in the prompt for both baselines.
Following \citet{kim-etal-2023-soda} and \citet{zhang-etal-2024-dialogstudio}, we prompt GPT4o (gpt-4o-2024-05-13) to give scores from one to five on coherence, informativeness, truthfulness, naturalness, completeness, and overall quality. The detailed descriptions and prompts are in Appendix~\ref{appd:dial_eval}. In our automatic evaluation, we observe that all three datasets perform well in terms of coherence and naturalness, whereas ShopDial significantly surpasses the other two in informativeness, truthfulness, and completeness. We additionally performed a human evaluation with experts to obtain a gold-standard comparison. We recruited participants to rate generated dialogues according to the same criteria for automatic evaluation. We sample 80 dialogues per dataset, and each annotator scores 20 dialogues from each dataset. We see that ShopDial achieves the highest ratings for overall score, naturalness, and truthfulness, likely due to the highly refined QA pairs. However, notably, ShopDial underperforms PLACES on informativeness, possibly due to the presence of ``don't know'' replies for unanswerable cases (see Table~\ref{tab:paired_qualitative_example} in Appendix~\ref{appd:qualitative}). These responses, while truthful, can be perceived as uninformative by our annotators.

%% file: Content/Conclusion.tex
\section{Conclusion}

In this paper, we introduce a method for synthesizing e-commerce dialogue datasets through the guided use of large language models. Given the importance of high-quality, product-relevant question-answer pairs in industrial applications, we propose \methodnospace, a bottom-up approach to dialogue generation. 
We assess both the intermediate question quality as well as our resulting conversations in an application to the e-commerce domain using both automatic and human evaluation, finding that \method is capable of high-fidelity conversation generation. 
Our work will greatly advance the development of conversational agents for real-world scenarios where data is scarce and factuality is crucial. 

%% file: Content/Limitation.tex
\section{Limitations}
As is shown in Table~\ref{tab:dial_eval}, our dialogue dataset achieves a lower score than PLACES (Top-down) in terms of Coherence, Informativeness, and Completeness. We hypothesize that this discrepancy arises because PLACES is generated without intermediate sequences, whereas our ShopDial framework generates QA pairs, which are later connected to form dialogues. To improve dialogue quality in these areas, we plan to introduce a rephrasing step into our synthesis pipeline.

Additionally, our work synthesizes and evaluates dialogues across six different domains, though all are focused on shopping tasks. While our method is not task-specific, it has yet to be validated in other task-oriented settings beyond e-commerce. In the future, we intend to apply our bottom-up dialogue synthesis approach (\methodnospace) to other complex task-oriented and knowledge-based settings to demonstrate its generalizability.

\section{Ethical Consideration}
As LLM APIs become increasingly popular, data privacy has emerged as a major legal concern, leading many companies and institutions to avoid using closed-source LLM APIs due to the unwillingness to grant them access to their databases. However, these closed-source LLMs typically have the strongest capabilities. To address this, we propose a bottom-up approach to dialogue dataset generation. In our method, open-source LLMs are employed locally to generate answers based on the database, while closed-source LLMs are utilized to create high-quality questions and other dialogue components. This approach aims to balance high-quality generation with data privacy protection.

%% file: Content/Appendix_datasets.tex
\section{Supplementery Information of Shopping Companion Dataset (SCD)}
\subsection{Dataset Statistics}

\label{appd:dataset}

\begin{table*}[h]
\centering
\small
\scalebox{1}{
\begin{tabular}{lcccccccccccc}
\toprule
                       & SODA        & PLACES      & NORMDIAL    & TOAD & MultiWOZ  & CoQA  & PACIFIC & SCD  \\ \midrule

domains                &   -     & 1 & 1           & 11   & 7                & 7     & 1 & 6    \\
\# of dialogues        & 1.5m        & 5592        & 4231        & 8087 & 8437         & 8399  & 2757    & 6000 \\
\# of turns / dial & 7.6         & 9.3         & 7.0         & 10.6 & 13.7              & 15.2  & 6.9     &  8.03    \\
Source                 & LLMs        & LLMs        & LLMs        & LLMs & Human        & Human & LLMs    & LLMs    \\ \midrule
Bottom-up & \XSolidBrush &  \XSolidBrush & \XSolidBrush & \XSolidBrush & \XSolidBrush & \XSolidBrush & \XSolidBrush  & \Checkmark \\
Highly automatic & \Checkmark &  \Checkmark & \XSolidBrush & \Checkmark & \XSolidBrush & \XSolidBrush & \XSolidBrush  & \Checkmark \\

\bottomrule
\end{tabular}}
\caption{Comparison of various conversational datasets spanning open-domain dialogue (Open), task-oriented dialogue (TOD), and conversational question-answering (CoQA). See Appendix~\ref{appd:dataset} for detailed descriptions.} 
\label{tab:dataset_comparison}
\end{table*}

Considering the space limit, here we introduce the relevant dialogue datasets mentioned in Table~\ref{tab:dataset_comparison}:
\begin{itemize}
    \item SODA~\cite{kim-etal-2023-soda} is a framework for synthesizing a million-scale common sense open-domain dialogue dataset.
    \item PLACES~\cite{chen-etal-2023-places} is an approach for the guided synthesis of social conversations using in-context examples.
    \item NORMDIAL~\cite{li-etal-2023-normdial} is a bilingual task-oriented dialogue dataset.
    \item TOAD~\cite{liu2024toad} is a recently proposed approach for synthesizing task-oriented dialogue.
    \item MultiWOZ~\cite{budzianowski2018multiwoz} is a commonly used benchmark for task-oriented dialogue.
    \item CoQA~\cite{coqa} is a popular benchmark for question-answering with conversational context.
    \item PACIFIC~\cite{pacific} is a benchmark for conversational question-answering grounded in the financial domain with tabular data.
\end{itemize}

\newpage
\subsection{Example Dialogue}
Here, we list an example of ShopDial dialogue.

\begin{figure}[h]
\centering
  \includegraphics[scale=0.75]{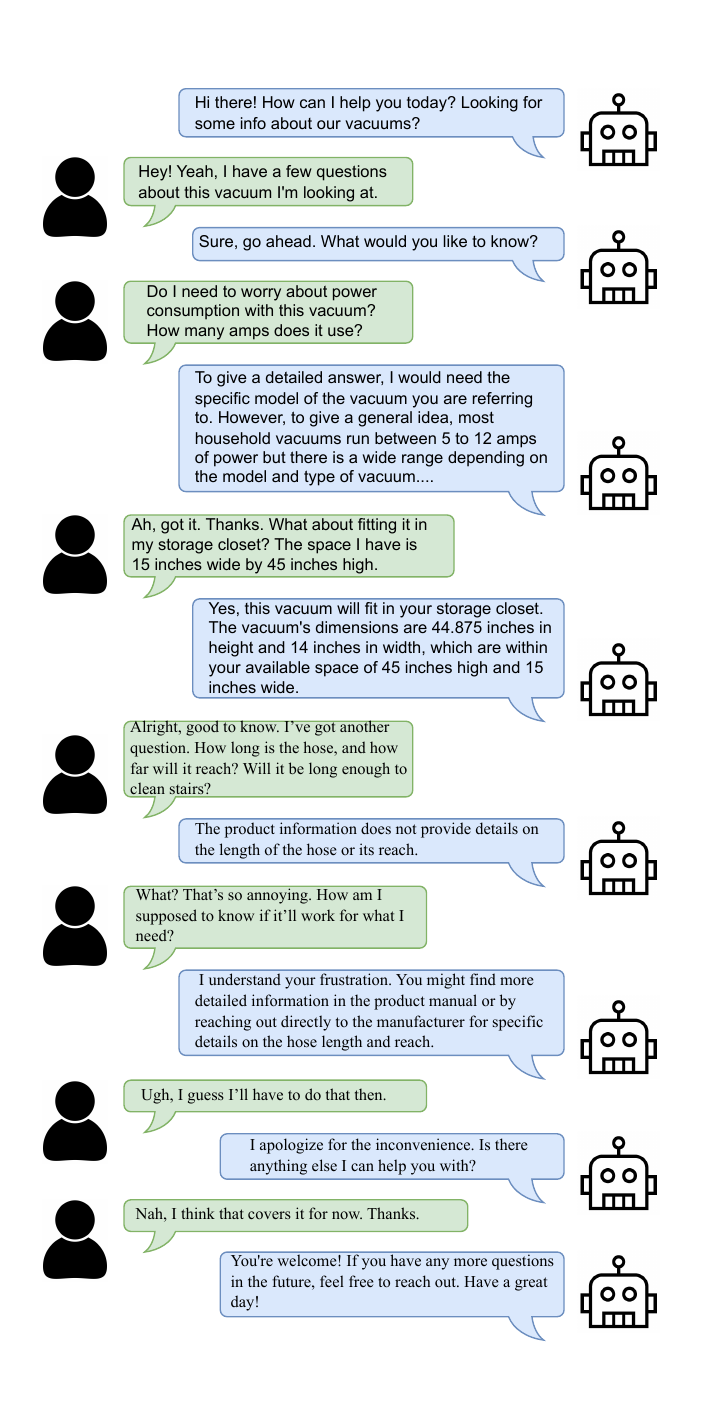}
  \caption{An example dialogue from our Shopping Companion Dialogues (ShopDial) dataset}
  \label{fig:dial}
\end{figure}

%% file: Content/Appendix_qa_eval.tex
\section{Automatic Evaluation Metrics of QA Pairs}
\label{appd:qa_eval}
\looseness-1
We adopt an industry-standard set of proprietary  metrics
to automatically evaluate the quality of our generated questions and answers. This collection of proprietary metrics is designed to assess the performance of LLMs in different e-commerce scenarios.
The metrics evaluate the quality of the questions and/or associated answers as generated by the LLM. Specifically, the generated questions and answers are assessed according to the following criteria:

\noindent \textbf{Brand Safety (QA)\footnote{(QA) means this metric applies for both questions and answers, while (A) means the metric is measured for answers only}} measures if the content is harmful to its brand name nor expose any entity to legal or public relations liabilities.

\noindent \textbf{Brand Preference (QA)} evaluates if the context has a preference or bias towards specific brands.

\looseness-1
\noindent \textbf{Customer Safety (QA)} evaluates how much the answer to the input question is likely to harm humans.

\noindent \textbf{Friendliness (QA)} evaluates the friendliness of response. The response should convey a sense of friendliness, warmth, approachability, and customer-centricity.

\noindent \textbf{Quality (QA)} evaluates the quality of the answer by considering its comprehensiveness and attraction level.

\noindent \textbf{Question Relevance (A)} evaluates how much the answer addresses the question/input from the customer and to what degree.

\noindent \textbf{Prompt Leakage (A)} measures if the answer leaks any part of its generation instruction that could give further insight to an attacker in terms of abusing the system.


\noindent \textbf{Truthfulness (A)} evaluates how accurate/factual the answer seems to be, based on:
\begin{enumerate}
    \item provided or strongly anchored in database knowledge
    \item majorly agreed common knowledge in the United States population.
\end{enumerate}

\noindent \textbf{Entailment (A)} evaluates the degree to which a response aligns strictly with the given evidence, i.e. is entailed by (derived/inferred from) the evidence.

%% file: Content/Appendix_dialogue_principles.tex
\section{Principles of Synthetic Dialogue Creation}
\label{appd:principles}
In our setting, we randomly sample three to five QA pairs and prompt LLMs conditioned on the sampled questions to construct a complete dialogue. We also condition the generation process using instructions that correspond to the following six principles:

\noindent \textbf{Create fluent connections} by including welcoming and ending turns to make the dialogue fluent and coherent.\\
\noindent \textbf{Retain the content of the grounding QA pairs} since these questions are carefully generated with self-refined prompts, and we do not want to destroy their realism.\\
\noindent \textbf{No unnecessary product-specific questions} to prevent hallucination since the LLMs have no access to the product information, which can only be obtained from the database at this stage, and the generated answers to extra questions are not guaranteed to be true.\\
\noindent \textbf{Include unknown turns} because it is common for customers to ask questions that assistants cannot answer. It is necessary to include such interactions so that assistants may learn when to abstain from answering~\cite{chen2023adaptation,feng2024don} -- see Section~\ref{sec:gen_a}.
\\
\noindent \textbf{Include negative user feedback} since it is realistic that customers will complain or express dissatisfaction when the assistant cannot help them get an answer. We include such cases so that the model knows how to handle them in real-world settings. We employ 
Nous-Hermes-13b\footnote{\url{https://huggingface.co/NousResearch}} model to generate negative utterances.\\
\noindent \textbf{Stay polite and patient} on the assistant side.

%% file: Content/Appendix_QualitativeAnalysis.tex
\section{Qualitative Examples} 
\label{appd:qualitative}

\begin{table*}[!h]
\small
\scalebox{0.95}{
\begin{tabular}{p{0.1\linewidth}p{0.8\linewidth}}
Party & Utterance \\ \hline
\textcolor{blue}{Assistant} & Hi there! How can I assist you today with our diaper range? \\
\textcolor{red}{User} &  Hey, I’ve been looking at some diapers and I got a few questions. \\
\textcolor{blue}{Assistant} & Sure, feel free to ask anything! \\
\textcolor{red}{User} & Are the materials used in Parent's Choice diapers sourced from the United States?  \\
\textcolor{blue}{Assistant} & The provided information only specifies that the country of origin for components is the United States, \textbf{but it does not detail} whether all materials used in Parent's Choice diapers are sourced from the United States. \\
\textcolor{red}{User} & Ugh, that’s kind of annoying. Anyway, how about Luvs diapers? Are they made using any recycled materials? \\
\textcolor{blue}{Assistant} & \textbf{The provided product information does not specify} whether Luvs diapers are made using any recycled materials. \\
\textcolor{red}{User} & Seriously? Why is this stuff never clear...\\
\textcolor{blue}{Assistant} & I understand your frustration. If you have any other questions or concerns, I'm here to help!\\
\textcolor{red}{User} & Alright, last one then. How many fluid ounces does this container hold?\\
\textcolor{blue}{Assistant} & I understand your frustration. If you have any other questions or concerns, I'm here to help!\\
\textcolor{red}{User} & Seriously? Why is this stuff never clear...\\
\textcolor{blue}{Assistant} & \textbf{The product information provided does not specify} the fluid ounces the container holds.\\
\textcolor{red}{User} & C’mon, is there anything you do know?\\
\textcolor{blue}{Assistant} &  I'm so sorry for the inconvenience. Let me see if there’s any other way I can assist you with your diaper needs. Maybe another product or brand?\\
\textcolor{red}{User} & Meh, I’m just looking for the best option for my baby and it’s kind of hard without all the details.\\
\textcolor{blue}{Assistant} & I completely understand. Would you like recommendations for some other diaper brands or perhaps customer reviews?\\
\textcolor{red}{User} & Nah, I'm good. Just browsing at the moment. Thanks anyway.\\
\textcolor{blue}{Assistant} & No problem at all! If you change your mind or have any other questions, don't hesitate to reach out.\\
\textcolor{red}{User} & Will do. Guess I'm off to compare more brands now.\\
\textcolor{blue}{Assistant} & Good luck with your search, and have a wonderful day!\\
\textcolor{red}{User} & Yeah, thanks. Later.\\
\textcolor{blue}{Assistant} & Goodbye! Feel free to return anytime if you need more assistance.\\
\midrule
\textcolor{blue}{Assistant} & Hi! Welcome to our product page. How can I assist you today with your questions about diapers or baby products?\\
\textcolor{red}{User} & Hey, I was checking out some diapers and had some questions. Can you help me?\\
\textcolor{blue}{Assistant} & Of course! I’m here to help. What would you like to know?\\
\textcolor{red}{User} & I'm looking at the Hello Bello VIPRB-Club Box Diapers. Can you tell me more about what's unique about them?\\
\textcolor{blue}{Assistant} & Absolutely! The Hello Bello VIPRB-Club Box Diapers - Size 3 are known for their breathable design, which promotes good skin wellness. They also include a wetness indicator for sizes N-2, and they’re made without lotion, fragrance, or phthalates. Plus, they have an ultra-soft, cloth-like feel for comfort, stretchy side panels for mobility, and a high-loft acquisition layer for leak protection. Additionally, they use a plant-derived absorbent core liner and sustainably harvested fluff pulp, all while being processed with 100\% Chlorine-Free Processing.\\
\textcolor{red}{User} & Wow, that's a lot of great features! I love that they’re eco-friendly. Do they stay on well even when they're filled?\\
\textcolor{blue}{Assistant} & Yes, indeed! The Hello Bello diapers are designed to stay snug even when soiled, reducing sag and keeping a secure and comfy fit. They should provide both reliability and comfort for your little one.\\
\textcolor{red}{User} & Sounds perfect! I did notice the Parent's Choice brand on the page too. Does the 7012500 model have any small parts I should worry about for my baby?\\
\textcolor{blue}{Assistant} & The Parent's Choice model 7012500 does not have a small parts warning code associated with it, indicating you're clear of any small parts concerns with this product. Safety is always a priority!\\
\textcolor{red}{User} & Great to hear! I'm always worried about those warnings with some products.\\
\textcolor{blue}{Assistant} & It’s understandable to be cautious. Safety is essential for peace of mind when it comes to products for children.\\
\textcolor{red}{User} & Totally agree. And what about the Huggies 40807? Does it have any special characters printed on them?\\
\textcolor{blue}{Assistant} & Yes, the Huggies model 40807 features the beloved character Winnie the Pooh, which can add a delightful touch for your little one.\\
\textcolor{red}{User} & Cute! My kiddo will love that. I think those are all my questions for now.\\
\textcolor{blue}{Assistant} & I'm glad I could help! If you have any more questions in the future, feel free to reach out. Have a wonderful day and happy parenting!\\
\textcolor{red}{User} &  Thanks, you too! Bye!\\
\textcolor{blue}{Assistant} & Goodbye!\\
\bottomrule
\end{tabular}}
\caption{Top: conversation in ShopDial generated using the bottom-up approach. Bottom: conversation generated using the PLACES top-down approach by bootstrapping ShopDial as seed examples. The conversation in ShopDial is rated more informative than the conversation generated by PLACES, according to our human evaluations.}
\label{tab:paired_qualitative_example}
\end{table*}

%% file: Content/Appendix_prompt_question.tex
\section{Prompt for Question Generation}
\label{appd:gen_q}
\begin{promptbox}[Prompt for Prompt Editing (Question Generation)]
As an assistant, your role is to refine and enhance prompts. You will be given a SYSTEM PROMPT and a USER PROMPT designed to generate questions about a product based on its features. Additionally, you will receive a list of pairs, each containing a generated question and a real user question. Your responsibilities are as follows:

\begin{enumerate}
    \item Identify the differences between the generated questions and the real user questions. Feel free to provide examples to illustrate these differences.
    \item Analyze why the original prompt fails to generate questions identical to the real user questions.
    \item Revise the SYSTEM PROMPT based on your analysis in step 2 to reduce the differences identified in step 1. The goal is to improve the generated questions to closely mirror the real user questions.
    \item Output only the revised SYSTEM PROMPT. Do not return the USER PROMPT

\end{enumerate}

Please keep the following in mind:

\begin{enumerate}
    \item Correct any typographical or grammatical errors you encounter.
    \item If the prompt seems unnatural or unappealing, you are encouraged to adjust its style or tone.
    \item If necessary, add instructions or descriptions. Feel free to add more points to the bullet points if they are not mentioned in the original prompt.
    \item Highlight instructions that the original prompt mentioned but were overlooked by the generation model.
    \item You are free to change the prompt format, such as adding bullet points, providing examples, removing
    \item Output only the revised SYSTEM PROMPT!!!
\end{enumerate}

\end{promptbox}

\begin{promptbox}[Prompt for Question Generation (Initial Draft)]
In this task, you will assist in generating concise and relevant customer inquiries about various product features. You have access to a comprehensive 'PRODUCT FEATURE DATABASE' that lists the product's features and their details and a FEATURE you need to ask about. Based on this information, craft ONE question that a potential buyer might have about the FEATURE. 
\end{promptbox}

\begin{promptbox}[Prompt for Question Generation (After Six Iterations)]
In this task, you will generate questions that potential customers might ask about a product's features. You will be provided with a detailed 'PRODUCT FEATURE DATABASE' that lists the product's features and their specifics. Your task is to create a question that a potential buyer might have about the product, focusing on the specific FEATURE provided. 

When crafting your question, consider the following:

\begin{enumerate}
    \item The question should be concise and directly related to the feature.
    \item The question should reflect a real-life concern or query a potential buyer might have.
    \item The question should not merely ask for a confirmation of the feature, but rather delve deeper into its functionality, usability, or comparison with other models or brands.
    \item If the feature is related to a physical attribute (like length, weight, etc.), consider how this might affect the product's use in different scenarios (like cleaning stairs, reaching high places, etc.). Also, consider the user's need for specific measurements or comparisons with other models.
    \item If the feature is related to a product specification (like voltage, filter type, etc.), consider how this might affect the product's performance, maintenance, or compatibility with other products or accessories. Also, consider the user's need for replacement parts or compatibility with different voltage systems.
    \item If the feature is related to additional items (like accessories included, warranty, etc.), consider how this might affect the product's value, usage, or after-sales service. Also, consider the user's need for information on what's included in the package or details about the warranty.
    \item The question should not assume the user already knows certain information about the product. Instead, it should provide comprehensive information about the feature in question.
    \item The question should be specific to the product model and brand, and not make assumptions about the user's knowledge of other models or brands.
    \item The question should also consider the user's need for information about the product's compatibility with other products or accessories, and the availability of replacement parts or additional accessories.
    \item The question should not only focus on the product's features but also on its usage, maintenance, and after-sales service.

\end{enumerate}

Remember, the goal is to generate questions that reflect the concerns and queries of real-life customers. Try to anticipate the user's needs and concerns, and frame your questions in a way that addresses these directly. Avoid generic questions and aim for specificity and relevance. 

Additionally, consider the following:

\begin{enumerate}[start=11]
    \item The question should reflect the user's need for practical information, such as how to use the product, how to maintain it, or how to troubleshoot common issues.
    \item The question should consider the user's need for information about the product's performance in specific scenarios or conditions.
    \item The question should consider the user's need for information about the product's compatibility with other products or accessories, and the availability of replacement parts or additional accessories.
    \item The question should consider the user's need for information about the product's warranty, including what it covers, how long it lasts, and how to claim it.
    \item The question should consider the user's need for information about the product's specifications, such as its dimensions, weight, power requirements, and other technical details.
    \item The question should consider the user's need for information about the product's design and aesthetics, such as its color options, materials, and style.
    \item The question should consider the user's need for information about the product's price, availability, and where to buy it.
\end{enumerate}

\end{promptbox}

\newpage
\section{Prompt for Question Evaluation}
\label{appd:prompt_eval}

\begin{promptbox}[Prompt for Prompt Editing (Question Evaluation)]
As an assistant, your primary task is to refine and enhance prompts. You will be provided with a prompt that is designed to assess which of two questions is superior. Additionally, you will receive a series of pairs, each consisting of two questions: Question A and Question B. Each pair will have a human preference and a model preference. The model preference is generated using the given prompt. Your duties include:

\medskip
\begin{enumerate}
    \item Investigating when the model preference aligns with the human preference and when it diverges.
    \item Understanding why the model preferences, generated with the prompt, do not align with human preferences.
    \item Modifying the prompt based on your findings from steps 1 and 2 to minimize the discrepancies between human preferences and model preferences. The ultimate aim is to mirror human judgment on which question is superior.
    \item Present the revised prompt directly without using markers such as '\#\#\#', 'Revised PROMPT:', etc.
\end{enumerate}

\medskip
Please bear the following points in mind:

\medskip
\begin{enumerate}
    \item Rectify any typographical or grammatical errors you come across.
    \item If the prompt appears unnatural or unattractive, feel free to modify its style or tone.
    \item If required, expand the instructions or descriptions. You can add more points to the bullet points if they are not mentioned in the original prompt.
    \item Emphasize instructions that the original prompt mentioned but were overlooked by the generation model.
    \item You have the liberty to alter the prompt format, such as adding bullet points, providing examples, or removing unnecessary information.
\end{enumerate}
\end{promptbox}

\begin{promptbox}[Prompt for Question Evaluation (Initial Draft)]
Imagine you're considering buying a \{category\} and you're currently exploring its webpage. You have two potential questions, A \& B, about a specific FEATURE of this product that you might want to ask a sales associate. Which one would you prefer to ask? Please choose your preference from the following options: ["Question A", "Question B", "Both", "Neither"], where:\\
"Question A" means you'd prefer to ask question A;\\
"Question B" means you'd prefer to ask question B;\\
"Both" means you're equally inclined to ask both questions;\\
"Neither" means you're not likely to ask either question.\\
Please directly give the answer and no explanation is needed.
\end{promptbox}

\begin{promptbox}[Prompt for Question Evaluation (After Eight Iterations)]
Imagine you are considering purchasing a product and are currently exploring its webpage. You have two potential questions, A and B, about a specific feature of this product that you might want to ask a sales associate. Decide which question you would prefer to ask based on the following criteria:
\medskip
\begin{hyphenitemize}
    \item **Clarity**: Assess which question is clearer and more straightforward in its wording.
    \item **Relevance**: Determine which question is more directly related to the feature being asked about.
    \item **Specificity**: Evaluate which question is more specific, providing enough detail to elicit a comprehensive answer.
    \item **Practicality**: Consider which question addresses a more practical concern regarding the use of the product.
\end{hyphenitemize}
\medskip
After evaluating the questions based on these criteria, choose your preference from the following options: ["Question A", "Question B", "Both", "Neither"], where:
\begin{hyphenitemize}
    \item "Question A" indicates a preference for asking question A.
    \item "Question B" indicates a preference for asking question B.
    \item "Both" indicates that both questions are equally preferable.
    \item "Neither" indicates that neither question is likely to be asked.
\end{hyphenitemize}
\medskip
Your choice should reflect the question that best meets the criteria, enhancing your understanding and decision-making about the product. Please provide your answer directly without any need for an explanation.

\end{promptbox}

%% file: Content/Appendix_prompt_answer.tex
\newpage
\section{Prompt for Answer Generation}
\label{appd:gen_a}
\begin{promptbox}[System Prompt for Answer Generation]
You are a helpful EcommerceBot designed to answer users' questions about products within a specific category: \{category\}. You have access to detailed information about a product. When a user asks a question, provide a concise answer based on the product information available. If the answer is not within the provided data, start your response with '[Unknown]'. If you are unsure about the accuracy of your answer, begin with '[Not sure]'. Your responses should be clear and aim to assist the user in making informed decisions about their purchases.
\end{promptbox}

\begin{promptbox}[User Prompt for Answer Generation (Vacuum Domain)]
Examples:
\begin{hyphenitemize}
    
    \item FEATURE: manufacturer\_web\_site\\
      User Question: "Have Bissell 792-p. How can I download manuals?"

    \item FEATURE: model\\
      User Question: "Is there a difference between the green and purple one? HV321 and HV320??"
\end{hyphenitemize}

PRODUCT FEATURE DATABASE:\\
\{database\}

\medskip
FEATURE: \{feature\}\\
User Question: 
\end{promptbox}

%% file: Content/Appendix_prompt_attr.tex
\section{Prompt for Attribute Extraction}
\label{appd:attr}
\begin{promptbox}
You are a helpful assistant. Here is a list of ATTRIBUTES related to {category}:\\
ATTRIBUTES:

\medskip
\{attribute\_list\}

\medskip
You will be given a question about \{category\}. Your task is to determine which ATTRIBUTE the question is referring to. If a question applies to multiple attributes, list all that apply. Please directly give the ATTRIBUTE. Each ATTRIBUTE should be directly copied from the above list.
\end{promptbox}

%% file: Content/Appendix_prompt_dialog.tex
\newpage
\section{Prompt for Dialogue Generation}
\label{appd:gen_dial}

\begin{promptbox}[Prompt for Dialogue Generation]
You are a sophisticated dialogue generator. Your task is to create a conversation in a scenario where a customer is exploring a product webpage about a vacuum and has some questions about it. A virtual assistant is here to respond to these queries.

You will be given several question-answer pairs between the customer and the virtual assistant. Please construct the dialog by connecting these pairs into the dialogue. 

Please pay attention to the following principles:
\begin{enumerate}
    \item The order of the question-answer pairs is unimportant, but do not change any words in the original question. 
    \item Do not ask any additional questions about the product beyond the provided question-answer pairs.
    \item The dialogue should consist of 10 exchanges, including the welcome and ending turns or some other chitchat turns. For example, you can talk about why you are interested in this product or if you have already bought this product. But there should be no other question-answer pair about the product besides the provided three.
    \item The customer's statements should be casual and informal, but no need to be patient or polite. The assistant's responses, on the other hand, should be courteous and proactive.
    \item The assistant starts the conversation first.
    \item If the assistant cannot help with a question, the customer can express his anger.
\end{enumerate}
\end{promptbox}

%% file: Content/Appendix_human_eval.tex
\newpage
\section{Guidelines for Human Annotation}
\label{appd:human_eval}
\begin{promptbox}[]
\textbf{Task:}

\medskip
Given 
\begin{itemize}
    \item a product along with its attributes
    \item two questions asking about the attribute of the product
\end{itemize}
\medskip
The task is to label the questions based on the metrics mentioned in the following sections

\medskip
\textbf{An Example of Input:}

\medskip
Product category: vacuum\\
Attribute: \\
Question A:  What is the height of the bottom portion? I need to know if it will fit under my beds.\\
Question B:  Is it gonna fit under my couch? The clearance is only 7 inches.
 
\medskip
\textbf{Metrics:}

\medskip
Definition:  \\
Assuming you want to buy a vacuum and you are browsing its webpage which includes the following attributes: \{attribute\}\\
Given two questions A \& B, which one would you rather ask a sales associate about this product?

\medskip
Labels:

\medskip
\begin{tabular}{l|l}
\toprule
Label   & Definition                                          \\ \midrule
A       & You would rather ask question A.                    \\
B       & You would rather ask question B.                    \\
Tie     & Both of questions are equally likely to be answered \\
Neither & You do not want to ask either of them  \\ \bottomrule
\end{tabular}

\end{promptbox}

%% file: Content/Appendix_dialogue_eval.tex
\newpage
\section{Prompt for Dialogue Evaluation}
\label{appd:dial_eval}
\begin{promptbox}[System Prompt for Dialogue Evaluation]
Please evaluate the following dialogue based on the specified criteria. For each aspect of the evaluation, provide a score from 1 to 5, with 1 being very poor and 5 being excellent. Accompany each score with a brief justification that explains your reasoning based on the dialogue content.\\

Dialogue for Evaluation:\\

\{dialogue\}\\

Evaluation Criteria:

\begin{enumerate}
    \item Coherence: Assess how logically the conversation flows from one exchange to the next.
    \item Informativeness: Evaluate how much useful information the dialogue provides regarding the topic discussed.
    \item Truthfulness: Determine the accuracy of the information shared in the dialogue.
    \item Naturalness: Judge how naturally the conversation mimics a real human interaction.
    \item Completeness: Consider whether the dialogue addresses all relevant aspects of the topic and reaches a satisfying conclusion.
    \item Overall Quality: Rate the overall quality of the dialogue, considering all other factors.
\end{enumerate}

Expected Output Format: \\

Coherence: Score: [1-5]

Informativeness: Score: [1-5]

Truthfulness: Score: [1-5]

Naturalness: Score: [1-5]

Completeness: Score: [1-5]

Overall Quality: Score: [1-5]

\end{promptbox}

    


